\documentclass[aip, floatfix, jcp, reprint, longbibliography, nofootinbib]{revtex4-1}

\usepackage[utf8]{inputenc}
\usepackage{amsmath}
\usepackage{amssymb}
\usepackage[nopar]{lipsum}
\usepackage{mathtools}
\usepackage{nameref,hyperref}
\usepackage{microtype}
\usepackage{color}
\usepackage{graphicx}
\usepackage{soul}
\usepackage{appendix}
\usepackage{hyperref}

\usepackage{alphalph}
\makeatletter
\newcommand*{\fnsymbolsingle}[1]{%
  \ensuremath{%
    \ifcase#1%
    \or *%
    \or \dagger
    \or \ddagger
    \or \mathsection
    \or \mathparagraph
    \else
      \@ctrerr
    \fi
  }%
}
\makeatother
\newalphalph{\fnsymbolmult}[mult]{\fnsymbolsingle}{}

\begin{document}

\title{PotentialNet for Molecular Property Prediction}

\author{Evan N. Feinberg*}
\email{enf@stanford.edu}
\affiliation{Program in Biophysics, Stanford University, Stanford CA 94305, USA}
\author{Debnil Sur}
\affiliation{Department of Computer Science, Stanford University, Stanford CA 94305, USA}
\author{Zhenqin Wu}
\affiliation{Department of Chemistry, Stanford University, Stanford CA 94305, USA}
\author{Brooke E. Husic}
\affiliation{Department of Chemistry, Stanford University, Stanford CA 94305, USA}
\author{Huanghao Mai}
\affiliation{Department of Computer Science, Stanford University, Stanford CA 94305, USA}
\author{Yang Li}
\affiliation{School of Mathematical Sciences and College of Life Sciences, Nankai University, China}
\author{Saisai Sun}
\affiliation{School of Mathematical Sciences and College of Life Sciences, Nankai University, China}
\author{Jianyi Yang}
\affiliation{School of Mathematical Sciences and College of Life Sciences, Nankai University, China}
\author{Bharath Ramsundar}
\affiliation{Department of Computer Science, Stanford University, Stanford CA 94305, USA}
\author{Vijay S. Pande*}
\email{pande@stanford.edu}
\affiliation{Department of Bioengineering, Stanford University, Stanford CA 94305, USA}

% \begin{abstract}
% \end{abstract}

\maketitle

\section*{Abstract}

The arc of drug discovery entails a multiparameter optimization problem spanning vast length scales. They key parameters range from solubility (angstroms) to protein-ligand binding (nanometers) to \textit{in vivo} toxicity (meters). Through feature learning---instead of feature engineering---deep neural networks promise to outperform both traditional physics-based and knowledge-based machine learning models for predicting molecular properties pertinent to drug discovery.  To this end, we present the PotentialNet family of graph convolutions. These models are specifically designed for and achieve state-of-the-art performance for protein-ligand binding affinity. We further validate these deep neural networks by setting new standards of performance in several ligand-based tasks. In parallel, we introduce a new metric, the Regression Enrichment Factor $EF_\chi^{(R)}$, to measure the early enrichment of computational models for chemical data. Finally, we introduce a cross-validation strategy based on structural homology clustering that can more accurately measure model generalizability, which crucially distinguishes the aims of machine learning for drug discovery from standard machine learning tasks.	

\section{Introduction}

Most FDA-approved drugs are small organic molecules that elicit a therapeutic response by binding to a target biological macromolecule. Once bound, small molecule ligands either inhibit the binding of other ligands or allosterically adjust the target's conformational ensemble. Binding is thus crucial to any behavior of a therapeutic ligand. To maximize a molecule's therapeutic effect, its affinity---or binding free energy ($\Delta G$)---for the desired targets must be maximized, while simultaneously minimizing its affinity for other macromolecules.
Historically, scientists have used both cheminformatic and structure-based approaches to model ligands and their targets, and most machine learning (ML) approaches use domain expertise-driven features. 

More recently, deep neural networks (DNNs) have been translated to the molecular sciences. Training most conventional DNN architectures requires vast amounts of data:~for example, ImageNet~\cite{deng2009imagenet} currently contains over $14,000,000$ labeled images. In contrast, the largest publicly available datasets for the properties of drug-like molecules include PDBBind 2017~\cite{liu2017forging}, with a little over $4,000$ samples of protein-ligand co-crystal structures and associated binding affinity values; Tox21 with nearly $10,000$ small molecules and associated toxicity endpoints; QM8 with around $22,000$ small molecules and associated electronic properties; and ESOL with a little over $1,000$ small molecules and associated solubility values\cite{wu2018moleculenet}. This scarcity of high-quality scientific data necessitates innovative neural architectures for molecular machine learning.

Successful DNNs often exploit relevant structure in data, such as pixel proximity in images.
Predicting protein-ligand binding affinity seems to resemble computer vision problems. Just as neighboring pixels connote closeness between physical objects, a binding pocket could be divided into a voxel grid. Here, neighboring voxels denote neighboring atoms and blocks of empty space. Unfortunately, this 3D convolutional approach has several potential drawbacks. First, inputs and hidden weights require much more memory in three dimensions. Second, since the parameters grow exponentially with the number of dimensions, the model suffers from the ``curse of dimensionality"~\cite{hastie2009overview}:~while image processing may entail a square $3^2$ filter, the corresponding filter for volumetric molecule processing has $3^3$ parameters.

In contrast, graph convolutions use fewer parameters by exploiting molecular structure and symmetry.
Consider a carbon bonded to four other atoms. A 3D CNN would need several different filters to accommodate the subgroup's symmetrically equivalent orientations. But a graph convolution as described in Refs.~\onlinecite{kearnes2016molecular, kipf2016semi, li2015gated, gilmer2017neural} is symmetric to permutations and relative location of each of the four neighbors, thereby significantly reducing the number of model parameters.
The use of graph convolutional approaches to learn molecular properties is reminiscent of the familiar canon of chemoinformatics algorithms such as Morgan fingerprints~\cite{morgan1965generation}, SMILES strings~\cite{weininger1989smiles}, and the Ullman algorithm for substructure search~\cite{ullmann1976algorithm}, all of which enrich chemical descriptions by propagating information about neighboring atoms.

In this paper, we first review a subset of DNN architectures applicable to protein-ligand interaction.
Through the mathematical frameworks above, we contextualize our presentation of new models that 
generalize a graph convolution to include both intramolecular interactions and noncovalent interactions between different molecules.
First, we describe a staged gated graph neural network, which distinguishes the derivation of differentiable bonded atom types from the propagation of information between different molecules. Second, we describe a more flexible model based on a new update rule using both the distance from source to target atom and the target atom's feature map.
Our direct incorporation of target atom information into the message function increases signal in some protein-ligand binding affinity benchmarks.
Finally, we address a potential shortcoming of the standard benchmark in this space---namely, treating the PDBBind 2007 core set as a fixed test set---by choosing a cross-validation strategy that more closely resembles the reality of drug discovery.
Though more challenging, this benchmark may better reflect a given model's generalization capacity.

\section{Neural Network Architectures} \label{sec:nns}

First, we briefly review a subset of DNN architectures applicable to protein-ligand interaction in order to motivate the new models we present and test at the end of the paper.

\subsection{Ligand-based scoring models}
\subsubsection{Fully Connected Neural Networks}
The qualitatively simplest models for affinity prediction and related tasks incorporate only features of ligands and ignore the macromolecular target(s). Such a model could entail a fully connected neural network (FCNN), in which each molecule is represented by a flat vector 
$x$ containing $f_0$ features. Then, these features are updated through ``hidden'' layers $h$ by applying nonlinear activation functions.

The training data for such a network consists of a set of $N$ molecules, each represented by a vector of length $f_0$, which have a one-to-one correspondence with a set of $N$ affinity labels.
Domain-expertise driven flat vector features might include integer counts of different types of pre-determined functional groups (e.g., carboxylic acids, aromatic rings), polar or nonpolar atoms, and other ligand-based features.
Cheminformatic featurizations include Extended Circular Fingerprints (ECFP)~\cite{rogers2010extended}
and ROCS~\cite{hawkins2007comparison, kearnes2016rocs}. 

\subsubsection{Graph Convolutional Neural Networks}

\begin{figure}
\includegraphics[width=234.0pt]{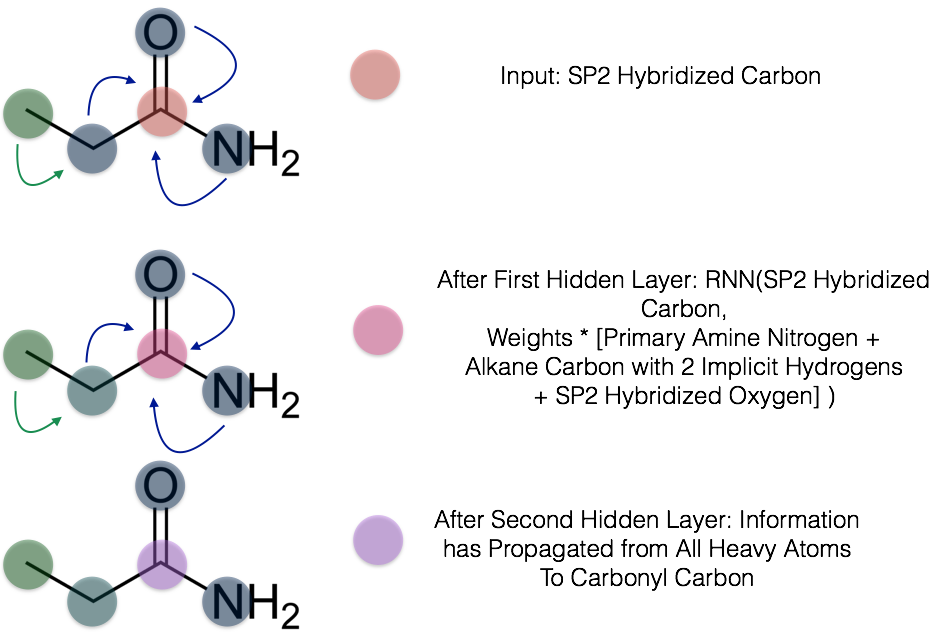}
\caption{Visual depiction of Gated Graph Neural Network with atoms as nodes and bonds as edges. The small molecule propanamide is chosen to illustrate the propagation of information among the different update layers of the network.}
\label{ggnn_fig}
\end{figure}

In convolutional neural networks (CNNs), each layer convolves the previous layer's feature map with linear kernels followed by elementwise nonlinearities, producing new features of higher complexity that combine information from neighboring pixels~\cite{krizhevsky2012imagenet}. A graph convolutional neural network (GCNN) analogously exploits the inherent structure of data ~\cite{duvenaud2015convolutional}.
We can represent a given graph that contains $N$ nodes, $f_{\text{in}}$ features per node, and a single edge type, as consisting of node features $x$ and symmetric adjacency matrix $A$, which designates whether a pair of nodes belong to each other's neighbor sets $N$. Consider the molecule propanamide (Figure \ref{ggnn_fig}). 
For the carbonyl carbon, the relevant row of the feature matrix $x$ might be ${[1, 0, 0]}$ to represent its element, and the corresponding row of the adjacency matrix $A$ might be ${[0, 1, 0, 1, 1]}$ to indicate its bonds to three neighbor atoms.

\noindent{} A graph convolution update, as summarized in ~\cite{gilmer2017neural}, entails applying a function at each node that takes the node and its neighbors as input and outputs a new set of features for each node. It can be written as

\begin{equation}\label{eq:mpnn}
h^{(t+1)}_i = U^{(t)} \left ( h^{(t)}_i, \sum_{v_j \in N(v_i)} m^{(t)} (h^{(t)}_j) \right ),
\end{equation}

\noindent{}where $h^{(t)}_i$ represents the node features of node $i$ at hidden layer $t$, $N(v_i)$ represents the neighbors of node $i$, and $U^{(t)}$ and $m^{(t)}$ are the update and message functions, respectively, at hidden layer $t$.
When there are multiple edge types, we must define multiple message functions, $m^{(t, e)}$, which is the message function at layer $t$ for edge type $e \in {[1, \dots, N_{\text{et}}]}$.

Our models are primarily inspired by
the Gated Graph Neural Networks (GGNNs)~\cite{li2015gated}. 
At all layers, the update function is the familiar gated recurrent unit (GRU). Message functions are simple linear operations that are different for each edge type but also the same across layers:

\begin{equation}\label{eq:symnorm}
h^{(t+1)}_i = GRU \left (h^{(t)}, \sum^{N_{\text{et}}}_e W^{(e)} A^{(e)} h^{(t)} \right ),
\end{equation}

\noindent{}where $A^{(e)}$ is the adjacency matrix, and $W^{(e)}$ the weight matrix, respectively, for edge type $e$. 

Unlike conventional FCNNs, which learn non-linear combinations of the input hand-crafted features, the update described in \eqref{eq:symnorm} learns nonlinear combinations of more basic features of a given atom with the features of its immediate neighbors. Information propagates through increasingly distant atoms with each graph convolution, and the GRU enables information to be added selectively. Ultimately, the GGNN contains and leverages both per-node features via the feature matrix $x$ and structural information via the adjacency matrix $A$.
In both classification and regression settings, GCNN's terminate in a ``graph gather'' step that sums over the rows of the final embeddings and is invariant to node ordering. The subsequent FCNNs produce output of desired size ($f_{\text{out}}$). 
This completes the starting point for the graph convolutional update used in this paper:

% \begin{widetext}
\begin{equation}
\begin{aligned}\label{GCNN_full}
 h^{(1)} &=& GRU \left (x, \sum^{N_{\text{et}}}_e W^{(e)} A^{(e)} x \right ) \\
 &\vdots& \\
 h^{(K)} &=& GRU \left (h^{(K-1)}, \sum^{N_{\text{et}}}_e W^{(e)} A^{(e)} h^{(K-1)} \right ) \\
 h^{(FC_0)} &=& \sum_{r=1}^N \left [ \sigma \left ( i(h^{(K)}, x) \right ) \odot \left (j(h^{(K)}) \right ) \right ]_r \\ 
 &\in& \mathbb{R}^{(1 \times f_{\text{out}})} \\
 h^{(FC_1)} &=& ReLU \left (W^{(FC_1)} \cdot h^{(FC_0)} \right ) \\
 & \vdots & \\
 h^{(FC_M)} &=& ReLU \left (W^{(FC_M)} \cdot h^{(FC_{M-1})} \right ).
\end{aligned}
\end{equation}
% \end{widetext}

\subsubsection{Generalization to multitask settings}
\noindent{}Predicting affinity for multiple targets by GCNN
can be implemented by training either different models for each target or by training a single multitask network. The latter setting has a last weight matrix $W^{(FC_M)} \in \mathbb{R}^{(T \times f_{FC_{M-1}})}$, where $T$ denotes the number of targets in the dataset. The corresponding multitask loss function would be the average binary cross-entropy loss across the targets, 

 \begin{equation}
 \begin{aligned}\label{multitask_loss}
 		\text{Loss}_{\text{multitask}} &=
\frac{1}{T} \sum^T_{j} \Bigg [ \frac{1}{n_j} \sum_i^{n_j} \Big( y_i \cdot \log(\sigma(\hat{y}_i)) \\ 
&+ (1 - y_i)\cdot 
         \log(1-\sigma(\hat{y}_i)) \Big) \Bigg ].
 	\end{aligned}
 \end{equation}

\subsection{Structure-based scoring models}
Since the advent of biomolecular crystallography by Perutz et al.~\cite{perutz1960structure}, the drug discovery community has sought to leverage structural information about the target in addition to the ligand. Numerous physics-based approaches have attempted to realize this, including molecular docking~\cite{jain2003surflex, shoichet1992molecular, trott2010autodock, friesner2004glide}, free energy perturbation~\cite{wang2015accurate}, and QM/MM~\cite{hensen2004combined}, among others. More recent approaches include RF-Score~\cite{li2015improving, ballester2010machine}, NN-score~\cite{durrant2011nnscore}, Grid Featurizer~\cite{wu2018moleculenet}, three dimensional CNN approaches~\cite{wallach2015atomnet, ragoza2017protein}, and Atomic Convolutional Neural Networks~\cite{gomes2017atomic}.

\subsubsection{PotentialNet Architectures for Molecular Property Prediction} \label{subsubsec:potentialnet}

To motivate architectures for more principled DNN predictors, we invoke the following notation and framework. 
First, we introduce the distance matrix $R \in \mathbb{R}^{(N \times N)}$, whose entries $R_{ij}$ denote the distance between $atom_i$ and $atom_j$. 
Thus far, the concept of adjacency, as encoded in a symmetric matrix $A$, has been restricted to chemical bonds. However, adjacency can also encompass a wider range of neighbor types to include non-covalent interactions (e.g., $\pi-\pi$ stacking, hydrogen bonds, hydrophobic contact).
Adjacency need not require domain expertise:~pairwise distances below a threshold value can also be used.
Regardless of particular scheme, we see how the distance matrix $R$ motivates the construction of an expanded version of $A$. In this framework, $A$ becomes a tensor of shape $N \times N \times N_{\text{et}}$, where $N_{\text{et}}$ represents the number of edge types.

If we order the rows by the membership of $atom_i$ to either the protein or ligand, we can view both $A$ and $R$ as block matrices, where the diagonal blocks are self-edges (i.e., bonds and non-covalent interactions) from one ligand atom to another ligand atom or from one protein atom to another protein atom, whereas off-diagonal block matrices encode edges from the protein to the ligand and from ligand to protein. For illustration purposes, we choose the special case where there is only one edge type, $N_{\text{et}}=1$:

\begin{equation}\label{blocksA}
	\begin{aligned}
	A &=
    \begin{bmatrix}
  A_{11} & A_{12} & \cdots & A_{1N} \\
  A_{21} & A_{22} & \cdots & A_{2N} \\
  \vdots & \vdots & \ddots & \vdots \\
  A_{N1} & A_{N2} & \cdots & A_{NN}
  \end{bmatrix}
			&= \begin{bmatrix}
			A_{L:L} & A_{L:P} \\
			A_{P:L} & A_{P:P} \end{bmatrix},
\end{aligned}
\end{equation}

\noindent{}where $A_{ij}$ is 1 for neighbors and 0 otherwise, and $A \in \mathbb{R}^{N \times N}$. Within this framework, we can mathematically express a \textbf{spatial graph convolution}---a graph convolution based on notions of adjacency predicated on Euclidean distance---as a generalization of the GGNN characterized by the update~\eqref{eq:symnorm}. 

In addition to edge type generalization, we introduce nonlinearity in the message portion of the graph convolutional layer: 

\begin{equation}
\begin{aligned}\label{introduce_nonlinearity}
 h^{(K)}_i &=& GRU \left (h^{(K-1)}_i, \sum^{N_{\text{et}}}_e \sum_{j \in N^{(e)}(v_i)} NN^{(e)} (h^{(K-1)}_j) \right ) ,
 \end{aligned}
 \end{equation}
 
\noindent{}where $NN^{(e)}$ is a neural network for each edge type $e$ and $N^{(e)}(h_i)$ denotes the neighbors of edge type $e$ for atom/node $i$. 
 
Finally, we generalize the concept of a layer to the notion of a \textbf{stage} that can span several layers of a given type. The Staged PotentialNet consists of three main steps: (1) covalent-only propagation, (2) dual non-covalent and covalent propagation, and (3) ligand-based graph gather. Stage (1), covalent propagation, entails only the first slice of the adjacency matrix, $A^{(1)}$, which contains a $1$ at entry $(i, j)$ if there is a bond between $(atom_i, atom_j)$ and a $0$ otherwise. Intuitively, stage (1) computes a new set of vector-valued atom types $h^{(1)}_i$ for each of the $N$ atoms in the system based on their local networks of bonded atoms. Subsequently, stage (2) entails propagation based on both the full adjacency tensor $A$ which begins with the vector-valued atom types $h^{(1)}_i$ computed in (1). While stage (1) computes new bond-based ``atom types'' for both amino acid and ligand atoms, stage (2) passes both bond and spatial information between the atoms. For instance, if stage (1) distinguishes an amide carbonyl oxygen from a ketone carbonyl oxygen, stage (2) might communicate in the first layer that that carbonyl oxygen is also within $3$ Angstroms of a hydrogen bond donor. Finally, in stage (3) a graph gather is performed solely on the ligand atoms. The ligand-only graph gather is made computationally straightforward by the block matrix formulation described in \eqref{blocksA}.

% \begin{widetext}
\textbf{PotentialNet, Stage 1:}
\begin{equation}\label{S-GGNN}
\begin{aligned}
 h^{(b_1)}_i &= GRU \left (x_i, \sum^{N_{\text{et}}}_e \sum_{j \in N^{(e)}(v_i)} NN^{(e)}  (x_j) \right ) \\ 
 &\vdots& \\
 h^{(b_K)}_i &= GRU \left (h^{(b_{K-1})}_i, \sum^{N_{\text{et}}}_e \sum_{j \in N^{(e)}(v_i)} NN^{(e)}  (h^{(b_{K-1})}_j) \right ) \\
 h^{(b)} &= \sigma \left (i^{(b)}(h^{(b_K)}, x) \right ) \odot  \left (j^{(b)} (h^{(b_K)}) \right ) \\ 
 &\in \mathbb{R}^{(N \times f_{b})}
 \end{aligned}
\end{equation}

\textbf{PotentialNet, Stage 2:}
\begin{equation}\label{S-GGNN2}
\begin{aligned}
 h^{(sp_1)}_i &= GRU \left (h^{(b)}_i, \sum^{N_{\text{et}}}_e \sum_{j \in N^{(e)}(v_i)} NN^{(e)}  (h^{(b)}_j) \right ) \\ 
  &\vdots& \\
 h^{(sp_K)}_i &= GRU \left (h^{(sp_{K-1})}_i, \sum^{N_{\text{et}}}_e \sum_{j \in N^{(e)}(v_i)} NN^{(e)}  (h^{(sp_{K-1})}_j) \right ) \\
 h^{(sp)} &= \sigma \left (i^{(sp)}(h^{(sp_K)}, h^{(b)}) \right ) \odot  \left (j^{(sp)} (h^{(sp_K)}) \right ) \\
 &\in \mathbb{R}^{(N \times f_{b})}
 \end{aligned}
\end{equation}

\textbf{PotentialNet, Stage 3:}
\begin{equation}\label{S-GGNN3}
\begin{aligned}
h^{(FC_0)} &= \sum^{N_{Lig}}_{j=1} h^{(sp)}_j \\
h^{(FC_1)} &= ReLU \left ( W^{(FC_1)} h^{(FC_0)} \right ) \\
&\vdots& \\
h^{(FC_K)} &= W^{(FC_K)} h^{(FC_{K-1})},
\end{aligned}
\end{equation}
% \end{widetext}

\noindent{}where $i^{(b)}$, $j^{(b)}$, $i^{(sp)}$, $j^{(sp)}$ are \textbf{b}ond and \textbf{sp}atial neural networks, and $h_j^{(sp)}$ denotes the feature map for the $j^{\text{th}}$ atom at the end of stage 2. 
\

\begin{figure}
\includegraphics[width=234.0pt]{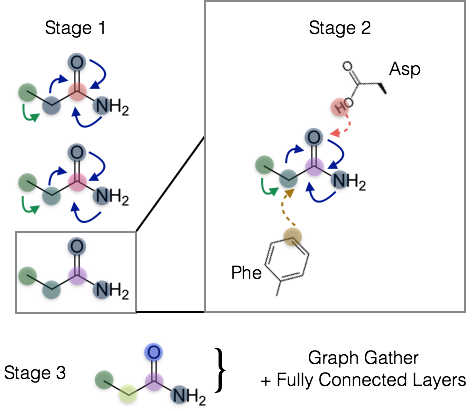}
\caption{Visual depiction of multi-staged spatial gated graph neural network. Stage 1 entails graph convolutions over only bonds, which derives new node (atom) feature maps roughly analogous to differentiable atom types in more traditional forms of molecular modeling.
Stage 2 entails both bond-based and spatial distance based propagation of information. In the final stage, a graph gather operation is conducted over the ligand atoms, whose feature maps are derived from bonded ligand information and spatial proximity to protein atoms.}
\label{sggnn_fig}
\end{figure}

A theoretically attractive concept in \eqref{S-GGNN} is that atom types---the $1 \times f_{b}$ per-atom feature maps---are derived from the same initial features for both ligand and protein atoms. In contrast to molecular dynamics force fields, e.g.~\cite{ponder2003force},
which---for historical reasons---have distinct force fields for ligands and for proteins which then must interoperate (often poorly) in simulation, our approach derives the physicochemical properties of biomolecular interactions from a unified framework.

To further illustrate, PotentialNet Stage 1 and Stage 2 exploit different subsets of the full adjacency tensor $A$:

\begin{figure}
\includegraphics[width=234.0pt]{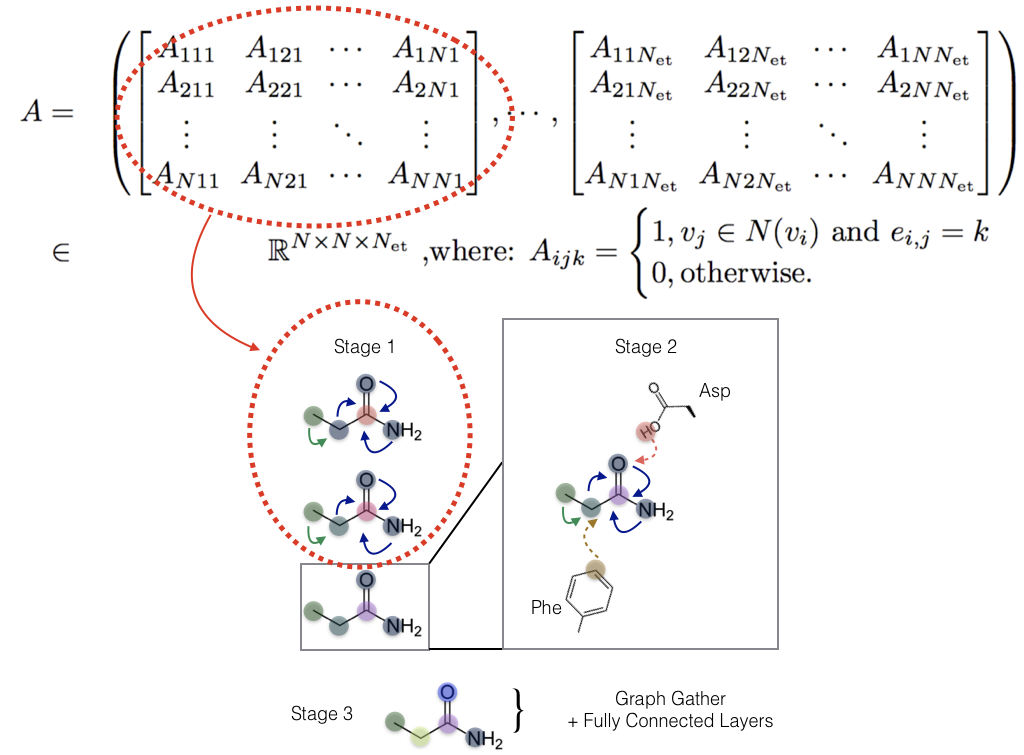}
\caption{PotentialNet Stage 1 exploits only covalent or bonded interaction edge types encoded in the first slices of the last dimension of the adjacency tensor $A$.}
\label{stage1}
\end{figure}

\begin{figure}
\includegraphics[width=234.0pt]{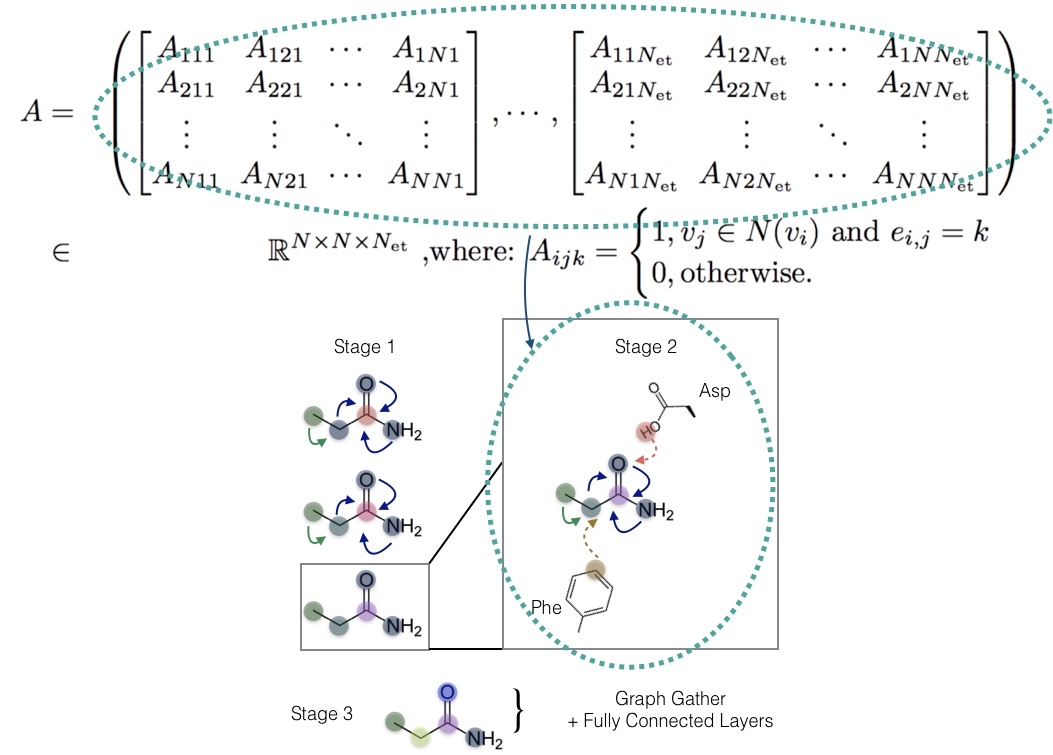}
\caption{PotentialNet Stage 2 exploits both bonded and non-bonded interaction edge types spanning the entirety of the last dimension of the adjacency tensor $A$.}
\label{stage2}
\end{figure}

\section{Measuring Early Enrichment in Regression Settings for Virtual Screening}
Traditional metrics of predictor performance suffer from general problems and drug discovery-specific issues. For regressors, both $R^2$---the ``coefficient of determination''---and the root-mean-square error ($RMSE$) are susceptible to single data point outliers. The $RMSE$ for both classifiers and regressors account for neither the training data distribution nor the null model performance. The area under the receiver operating characteristic curve $(AUC)$~\cite{triballeau2005virtual} does correct this deficiency in $RMSE$ for classifiers. However, all aforementioned metrics are global statistics that equally weight all data points. This property is particularly undesirable in drug discovery, which is most interested in predicting the tails of a distribution:~while
model predictions are made against an entire library containing millions of molecules, one will only purchase or synthesize the top scoring molecules.
In response, the cheminformatics community has adopted the concept of \textbf{early enrichment}. Methods like $BEDROC$~\cite{truchon2007evaluating} and $LogAUC$~\cite{mysinger2010rapid} weight the importance of the model's highest performers more heavily.

\subsection{Proposed Metric: $EF^{(R)}_\chi$}

At present, this progress in early enrichment measurement has been limited to classification and has yet to include regression. Therefore, we propose a new metric for early enrichment in regression, $EF^{(R)}_\chi$, analogous to $EF_\chi$. For a given target:

\begin{align}\label{efr_def}
EF^{(R)}_\chi = \frac{1}{\chi \cdot N} \sum^{\chi \cdot N}_{i} \frac{y_i - \bar{y}}{\sigma(y)} = \frac{1}{\chi \cdot N} \sum^{\chi \cdot N}_{i} z_i,
\end{align}

\noindent{}in which $y_i$, the experimental (observed) measurement for sample $i$, are ranked in descending order according to $\hat{y}_i$, the model (predicted) measurement for sample $i$. In other words, we compute the average z-score for the observed values of the top $\chi \%$ scoring samples. We prefer this approach to computing, for example, $\frac{1}{\chi \cdot N} \sum^{\chi \cdot N}_{i} \left (y_i - \bar{y}\right )$, which has units that are the same as $y_i$ (i.e., $log(K_i)$ values). Unfortunately, this unnormalized approach depends on the distribution in the dataset. For instance, in a distribution of $log(K_i)$ measurements, if the maximum deviation from the mean is $1.0$, the best a model can possibly perform would be to achieve an $EF^{(R)}_\chi$ of $1.0$.

We normalize through division by $\sigma(y)$, the standard deviation of the data. This allows comparison of model performance across datasets with a common unit of measurement but different variances in those measurements. 
The upper bound is therefore equal to the right hand side of~\eqref{efr_def}, where the indexed set of molecules $i$ constitutes the subset of the $\chi \cdot N$ most experimentally active molecules.
This value is dependent on both the distribution of the training data as well as the value $\chi$.
The $EF_\chi^{(R)}$ is an average over $\chi \cdot N$ z-scores, which themselves are real numbers of standard deviations away from the mean experimental activity.\footnote{
$EF_\chi^{(R)}$ values may therefore exceed 1.0,  since this means that the $\chi$ percentage of top predicted molecules have an average standard deviation of more than 1.0 above the mean.}

\section{Results}
\subsection{Cross-validation strategies}

It is well known that the performance of DNN algorithms is highly sensitive to chosen hyperparameters. Such sensitivity underscores the criticality of rigorous cross-validation~\cite{boulesteix2015ten, chicco2017ten}.  Several recent papers, including works that claim specifically to improve binding affinity prediction on the PDBBind dataset ~\cite{cang2017topologynet, jimenez2018k}, engage in the practice of searching hyperparmeters \textit{directly on the test set}. Compounding this problem is a fundamental deficiency of the main cross-validation procedure used in this subfield that is discussed below.

While there are newer iterations of the PDBBind dataset, e.g., Ref.~\onlinecite{liu2017forging}, we choose to evaluate performance on PDBBind 2007~\cite{wang2004pdbbind, wang2005pdbbind} to compare performance of our proposed architectures to previous methods. In previous works, the PDBBind 2007 dataset was split by (1) beginning with the ``refined'' set comprising $1,300$ protein-ligand co-crystal structures and associated binding free energy; (2) removing the ``core'' set comprising $195$ samples to form the test set, with (3) the remaining $1,095$ samples serving as the training data. We term this train-test split ``PDBBind 2007, Refined Train, Core Test'' below, and compare performance with RF-score~\cite{ballester2010machine}, X-Score~\cite{wang2002further, wang2003comparative}, and the networks \eqref{S-GGNN}-\eqref{S-GGNN3}  described in this work.

One drawback to train-test split is possible overfitting to the test set through hyperparameter searching. Another limitation is that train and test sets will contain similar examples. Whereas it is typical in other machine learning disciplines for the train and test set examples to be drawn from the same statistical distributions, such a setting is not necessarily desirable in a molecular machine learning setting~\cite{martin2017profile}. Drug discovery campaigns typically involve the synthesis and investigation of novel chemical matter. To accurately assess the generalizability of a trained model, the cross-validation strategy should reflect how that model will be deployed practically.
In context of this reasoning, 
the ``Refined Train, Core Test'' strategy is not optimal for cross-validation.
For example,~\citet{li2017structural}, showed that systematically removing samples from the PDBBind 2007 refined set with structural or sequence homology to the core (test) set significantly attenuated the performance of recent ML-based methods for affinity prediction.

\begin{figure}[h!]
\includegraphics[width=234.0pt]{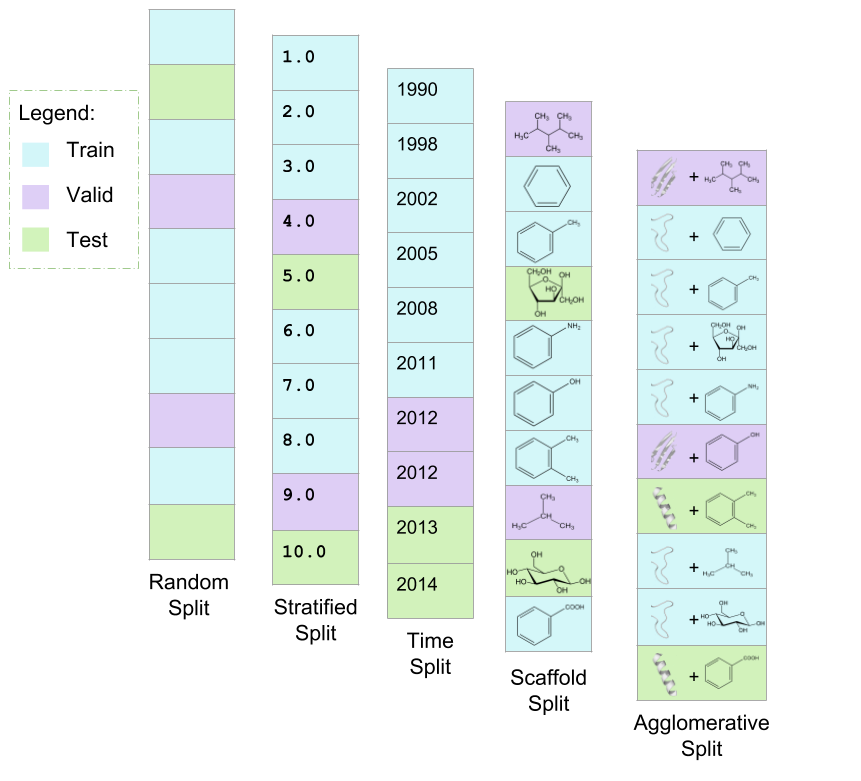}
\caption{Notional comparison of cross-validation splitting algorithms.
The first four vertical panels from the left depict simple examples of random split, stratified split, time split, and scaffold split. The rightmost panel depicts a toy example of the agglomerative split proposed in this work.
Both scaffold split and agglomerative split group similar data points together in order to promote the generalizability of the network to new data.
Scaffold split uses the algorithm introduced by \citet{bemis1996properties} to group ligands into common frameworks.
The agglomerative split uses hierarchical agglomerative clustering to group ligand-protein systems according to pairwise sequence or structural similarity of the proteins.
This figure is adapted from Ref.~\onlinecite{wu2018moleculenet} with permission from the Royal Society of Chemistry.}
\label{splitting_fig}
\end{figure}

Therefore, we propose and investigate a cross-validation strategy that splits all data into three distinct folds---train, validation, and test subsets---with agglomerative hierarchical clustering based on pairwise structural and sequence homology of the \textit{proteins} as distance metrics~\cite{husic2017unsupervised, kramer2010leave}.
Figure~\ref{splitting_fig} contrasts this technique with other common splitting methods for ligand binding studies.
Both sequence and structural similarity measures are described in Ref.~\onlinecite{li2017structural}.
The agglomerative clustering procedure is described in detail in Ref.~\onlinecite{husic2017unsupervised} and is a specific case of the method introduced in Ref.~\onlinecite{kramer2010leave}.
Our cross-validation on the PDBBind 2007 refined set with sequence similarity resulted in 978 train samples, 221 valid samples, and 101 test samples (75\%-17\%-8\%); meanwhile, clustering on structural similarity yielded 925 train samples, 257 valid samples, and 118 test samples (71\%-20\%-9\%).
A supplementary file is provided with the two sets of train, validation, and test assignments.

\subsection{Performance of methods on benchmarks}

\begin{table*}[t!]
	\caption{Benchmark:~PDBBind 2007, Refined Train, Core Test. Error bars are recorded as standard deviation of the test metric over three random initializations of the best model as determined by average validation set score. MUE is mean unsigned error. Pearson test scores for TopologyNet are reported from Ref.~\onlinecite{cang2017topologynet} and RF- and X-Scores are reported from Ref.~\onlinecite{li2017structural}.}
    \label{refinedtrain_coretest}
    \begin{tabular}{l l l l l l l l}
    \hline
    Model & Test $R^2$ & Test $EF_\chi^{(R)}$ & Test Pearson & Test Spearman & Test stdev & Test MUE \\ \hline
		PotentialNet & 0.668 (0.043) & 1.643 (0.127) & 0.822 (0.021) & \textbf{0.826} (0.020) & 1.388 (0.070) & 0.626 (0.037) \\ \hline
        PotentialNet, & 0.419 (0.234) & 1.404 (0.171)  & 0.650 (0.017) & 0.670 (0.014) & 1.832 (0.135) & 0.839 (0.005) \\ 
         (ligand-only control) & & & & & & \\ \hline
        TopologyNet, & N/A & N/A & \textbf{0.826} & N/A & N/A & N/A \\
        \textit{No Valid. Set} & & & & & & \\ \hline
    RF-Score & N/A & N/A & 0.783 & 0.769 & N/A & N/A \\ \hline
    X-Score & N/A & N/A & 0.643 & 0.707 & N/A & N/A \\ \hline
    \end{tabular}
\end{table*}

On the standard PDBBind 2007 ``refined train, core test'' benchmark, the PotentialNet Spatial Graph Convolution achieves state-of-the-art performance as reflected by several metrics. PotentialNet outperforms RF-Score and X-Score according to Pearson and Spearman correlation coefficients. The Pearson correlation score for \eqref{S-GGNN}-\eqref{S-GGNN3} is within error of the reported score for TopologyNet, the heretofore top performing model on this benchmark. However, a key caveat must be noted with respect to this comparison:~all
cross-validation for this manuscript, including all of our results reported in Tables \ref{refinedtrain_coretest}, \ref{sequence_split}, and \ref{structure_split}, was performed such that performance on the \textit{test} set was recorded for the hyperparameter set that performed most highly on the  \textit{validation} set. In contrast, the TopologyNet paper~\cite{cang2017topologynet}, models were trained on a \textit{combination} of the validation and training sets and evaluated \textit{directly on the test set}. Performance for TopologyNet~\cite{cang2017topologynet} therefore reflects a train-validation type split rather than a train-validation-test split, which likely inflated the performance of that method.

\begin{table*}
	\centering
	\caption{Benchmark:~PDBBind 2007 Refined, Agglomerative Sequence Split. Error bars are recorded as standard deviation of the test metric over three random initializations of the best model as determined by average validation set score. MUE is mean unsigned error. X-score does not have error because it is a deterministic linear model.}
    \label{sequence_split}
    \begin{tabular}{l l l l l l l l}
    \hline
    \textbf{Model} & Test $R^2$ & Test $EF_\chi^{(R)}$ & Test Pearson & Test Spearman & Test MUE \\ \hline
	PotentialNet & 0.480 (0.030) & 0.867 (0.036) & 0.700 (0.003) & 0.694 (0.012) & 1.680 (0.061) \\ \hline
        Ligand-only PotentialNet & 0.414 (0.058) & 0.883 (0.025) & 0.653 (0.031) & 0.674 (0.020) & 1.712 (0.110) \\ \hline 
    RF-score & \textbf{0.527} (0.014) & 1.078 (0.143) & \textbf{0.732} (0.009) & 0.723 (0.013) & \textbf{1.582} (0.034) \\ \hline
    X-score & 0.470  & \textbf{1.117} & 0.702 & \textbf{0.764} & 1.667 \\ \hline
    \end{tabular}
\end{table*}

Intriguingly, the gap in performance between the PotentialNet Spatial Graph Convolution and the other tested statistical models changes considerably on the agglomerative structure and sequence split benchmarks. On sequence split, RF-score achieves the best overall performance, followed by a statistical tie between the Staged Spatial Graph Convolution \eqref{S-GGNN}-\eqref{S-GGNN3} and X-Score, followed by the ligand-only graph convolutional control. Meanwhile, on structure split, PotentialNet achieves the highest overall performance, followed by RF-Score, followed by a statistical tie of X-Score, and the graph convolutional ligand-only control.

\begin{table*}
	\centering
	\caption{Benchmark:~PDBBind 2007 Refined, Agglomerative Structure Split. Error bars are recorded as standard deviation of the test metric over three random initializations of the best model as determined by average validation set score. MUE is mean unsigned error.  X-score does not have error because it is a deterministic linear model.}
    \label{structure_split}
    \begin{tabular}{l l l l l l l l}
    \hline
    \textbf{Model} & Test $R^2$ & Test $EF_\chi^{(R)}$ & Test Pearson & Test Spearman & Test MUE \\ \hline
	PotentialNet & \textbf{0.629} (0.044) & \textbf{1.576} (0.053) & \textbf{0.823} (0.023) & \textbf{0.805} (0.019) & \textbf{1.553} (0.125) \\ \hline
        Ligand-only PotentialNet & 0.500 (0.010) & 1.498 (0.411) & 0.733 (0.007) & 0.726 (0.005) & 1.700 (0.067) \\ \hline 
    RF-score & 0.594 (0.005) & 0.869 (0.090) & 0.779 (0.003) & 0.757 (0.005) & 1.542 (0.046) \\ \hline
    X-score & 0.517 & 0.891 & 0.730 & 0.751 & 1.751 \\ \hline
    \end{tabular}
\end{table*}

It is noteworthy that the PotentialNet Spatial Graph Convolutions  (\eqref{S-GGNN}-\eqref{S-GGNN3} perform competitively with other compared methods when the proposed Spatial Graph Convolutions are predicated on very simple, per-atom features and pure notions of distance whereas RF-Score, X-Score, and TopologyNet all directly incorporate domain-expertise driven information on protein-ligand interactions.

\subsubsection{Sanity check with a traditional RNN}

Given the unreasonable effectiveness of deep learning methods in mostly unstructured settings, it is important to justify our incorporation of domain knowledge over a purely deep learning-based approach. To do this, we trained a bidirectional long short-term memory (LSTM) network, a commonly-used recurrent neural network (RNN) that handles both past and future context well. We represented the protein-ligand complexes using a sequential representation of protein-ligand complexes in PDBBind:~proteins were one-hot encoded by amino acid, and ligands were similarly encoded on a character-level using their SMILES string representation. The test Pearson correlation coefficient corresponding to the best validation score (using the same metric) was 0.518, far worse than our results and justifying our model's incorporation of domain knowledge.

\subsection{Ligand-based models} \label{subsec:ligand}

While crystallography, NMR, and, most recently, cryo electron microscopy have opened a new paradigm of structure-based drug design, many critical tasks of drug discovery can be predicted from the chemical composition of a given molecule itself, without explicit knowledge of the macromolecule(s) to which they bind. Such properties include electronic spectra (important for parameterizing small molecule force fields for molecular dynamics simulations, for example), solubility, and animal toxicity.

Quantum mechanical datasets are particularly ripe for machine learning algorithms since it is straightforward to generate training data at some known accuracy. The QM8 dataset~\cite{ramakrishnan2015electronic}, which contains several electronic properties for small molecules in the GDB-8 set, lends itself particularly well for benchmarking PotentialNet \eqref{S-GGNN}-\eqref{S-GGNN3} since each compound's properties are calculated based on the three-dimensional coordinates of each element. The ESOL solubility~\cite{delaney2004esol} and Tox21 toxicity~\cite{tice2013improving} datasets map two-dimensional molecular representations consisting solely of atoms and their bonds to their respective single-task and multi-task outputs, and therefore serve as validation of our neural network implementations as well as of the value of incorporating nonlinearity into the message function. 

To summarize, our computational experiments indicate that PotentialNet leads to statistically significant improvements in performance for all three investigated ligand-based tasks. 
For the QM8 dataset, we were able to directly assess the performance benefit that stems from separating spatial graph convolutions into distinct stages.
Recall that Stage I of PotentialNet \eqref{S-GGNN} propagates information over only bonds and therefore derives differentiable ``atom types'', whereas Stage II of PotentialNet \eqref{S-GGNN2} propagates information over both bonds and different binned distances.
We performed an experiment with QM8 in which Stage I was essentially skipped, and graph convolutions propagated both covalent and non-covalent interactions without a privileged first stage for only covalent interactions. Clearly, separating the two stages led to a significant boost in performance. 

For each ligand model investigation we benchmark against the error suggested upon introduction of the dataset, or in order to enable direct comparison with previously published approaches. For extensive benchmarking of various models on these and other datasets, we refer the reader to Ref.~\onlinecite{wu2018moleculenet}.

\subsubsection{Quantum Property Prediction}

Table~\ref{table:qm8} reports the performances in mean absolute error (MAE) over 21786 compounds and 12 tasks in QM8.
We utilize MAE for consistency with the original proposal of the database~\cite{ramakrishnan2015electronic}.
Multiple PotentialNet variants and two mature deep learning models:~deep tensor neural network~\cite{schutt2016quantum} (DTNN) and message passing neural network~\cite{gilmer2017neural} (MPNN) are evaluated, in which the latter two models proved to be successful on quantum mechanical tasks (e.g.,~atomization energy\cite{wu2018moleculenet}). We restricted the training length to 100 epochs and performed 100 rounds of hyperparameter search on PotentialNet models. Staged spatial PotentialNet model achieved the best performances in the group, demonstrating strong predictive power on the tasks. We have also included taskwise results in Appendix~\ref{app:qm8}.

\begin{table*}
    \caption{Quantum Property Prediction with QM8 Dataset. Error bars are recorded as standard deviation of the test metric over three random initializations of the best model as determined by average validation set score.}
    \label{table:qm8}
    \begin{tabular}{l l l l}
    \hline
    Network & Valid MAE & Test MAE &  \\ \hline
        Spatial PotentialNet, Staged & 0.0120 & \textbf{0.0118} (0.0003)  \\ \hline
        Spatial PotentialNet, SingleUpdate & 0.0133 & 0.0131 (0.0001) \\ \hline
        MPNN & 0.0142 & 0.0139 (0.0007) \\ \hline
        DTNN & 0.0168 & 0.0163 (0.0010) \\ \hline
    \end{tabular}
\end{table*}

\subsubsection{Toxicity}

In the multitask toxicity benchmark, we evaluated the performances of two graph convolutional type models~\cite{kearnes2016molecular,duvenaud2015convolutional} and PotentialNet on the Tox21 dataset under same evaluation pattern. With 100 epochs of training, PotentialNet demonstrated higher ROC-AUC scores on both validation and test scores, outperforming Weave and GraphConv by a comfortable margin.

\begin{table*}
    \caption{Toxicity Prediction with the Tox21 Dataset. Error bars are recorded as standard deviation of the test metric over three random initializations of the best model as determined by average validation set score.}
    \label{table:tox21}
    \begin{tabular}{l l l l}
    \hline
    Network & Valid ROC AUC & Test ROC AUC &  \\ \hline
        PotentialNet & 0.878 & \textbf {0.857} (0.006) &  \\ \hline
        Weave & 0.852 & 0.831 (0.013) & \\ \hline
        GraphConv & 0.858 & 0.838 (0.001)  \\ \hline
        XGBoost & 0.778 & 0.808 (0.000) \\ \hline
    \end{tabular}
\end{table*}

\subsubsection{Solubility}

The same three models are also tested and compared on a solubility task~\cite{delaney2004esol}, using $RMSE$ to quantify the error in order to compare to previous work~\cite{duvenaud2015convolutional}. 
PotentialNet achieved slightly smaller $RMSE$ than Weave and GraphConv (Table~\ref{table:esol}). Under the limited 100 epochs training, the final test RMSE is comparable or even superior to the best scores reported for Weave and GraphConv (0.46~\cite{kearnes2016molecular} and 0.52~\cite{duvenaud2015convolutional} respectively).

\begin{table*}
    \caption{Solubility Prediction with the Delaney ESOL Dataset. Error bars are recorded as standard deviation of the test metric over three random initializations of the best model as determined by average validation set score.}
    \label{table:esol}
    \begin{tabular}{l l l l}
    \hline
    Network & Valid RMSE & Test RMSE \\ \hline
        PotentialNet & 0.517 & \textbf{0.490} (0.014) \\ \hline
        Weave & 0.549 & 0.553 (0.035) \\ \hline
        GraphConv & 0.721 & 0.648 (0.019)  \\ \hline
        XGBoost & 1.182 & 0.912 (0.000) \\ \hline
    \end{tabular}
\end{table*}

\begin{table*}
	\caption{Hyperparameters for neural networks \eqref{S-GGNN}-\eqref{S-GGNN3}.}
	\label{table:hyperparameters1}
    \begin{tabular}{l l l l}
    \hline
    Network & Hyperparameter Name & Symbol & Possible Values \\ \hline
        PotentialNet & Gather Widths (Bond and Spatial) & $f_{bond}, f_{spatial}$ & {[64, 128]} \\ \hline
        PotentialNet & Number of Bond Convolution Layers & $bond_K$ & {[1, 2]} \\ \hline
        PotentialNet & Number of Spatial Convolution Layers & $spatial_K$ & {[1, 2, 3]} \\ \hline
        PotentialNet & Gather Width & $f_{gather}$ & {[64, 128]} \\ \hline
        PotentialNet & Number of Graph Convolution Layers & $K$ & {[1, 2, 3]} \\ \hline
        Both & Fully Connected Widths & $n_{rows}$ of $W^{(FC_i)}$ & {[[128, 32, 1], [128, 1], [64, 32, 1], [64, 1]]} \\ \hline
        Both & Learning Rate & - & {[1e-3, 2e-4]} \\ \hline
        Both & Weight Decay & - & {[0., 1e-7, 1e-5, 1e-3]} \\ \hline
        Both & Dropout & - & {[0., 0.25, 0.4, 0.5]} \\ \hline
    \end{tabular}
\end{table*}

\section{Discussion}

Spatial Graph Convolutions exhibit state-of-the-art performance in affinity prediction. Whether based on linear regression, random forests, or other classes of DNNs, all three of RF-Score, X-Score, and TopologyNet are machine learning models that explicitly draw upon traditional physics-based features. Meanwhile, the Spatial Graph Convolutions presented here use a more principled deep learning approach. Input features are only basic information about atoms, bonds, and distances. This framework does not use traditional hand-crafted features, such as hydrophobic effects, $\pi$-stacking, or hydrogen bonding. Instead, higher-level interaction ``features" are learned through intermediate graph convolutional neural network layers. 

The traditional PDBBind 2007 benchmark uses $1,105$ samples from the refined set for training and $195$ from the core set for testing. Here, Spatial Graph Convolutions outperform X-Score and RF-Score and perform comparably with TopologyNet (even though this searched hyperparameters directly over the test dataset). On our proposed agglomerative clustering cross-validation benchmark, the choice of sequence or structure split affects relative performance. On sequence split, RF-Score achieved the highest overall performance, with Staged Spatial Graph Convolutions and X-Score statistically tied for second. But on structure split, the Staged Spatial Graph Convolutions performed best, with RF-score in second place. 

While the Pearson correlation was employed in the preceding performance comparison, instead comparing methods through $EF_\chi^{(R)}$ tells a mildly different story.
On the agglomerative sequence cross-validation split, in which test proteins are separated from train proteins based on amino acid sequence deviation, X-Score statistically ties RF-Score for the best model, while PotentialNet statistically ties the ligand-only PotentialNet control for last place at over $0.1$ average standard deviations worse than X-Score and RF-Score for the top $5\%$ of predictions.
Meanwhile, using the agglomerative structure cross-validation split, PotentialNet exceeds the performance of X-Score and RF-Score by over $0.5$ average standard deviations, though is within a statistical tie of the ligand-only PotentialNet control (which has a surprisingly high variance in its $EF_\chi^{(R)}$).
Taken together, we aver that it is important for the future of ML-driven structure-based drug discovery to carefully choose both (1)~the cross-validation technique and (2)~the metric of performance on held-out test set in order to most accurately reflect the capacity of their models to generalize in simulated realistic settings. 

In light of the continued importance and success of ligand-based methods in drug discovery, we benchmarked PotentialNet on several ligand based tasks: electronic property (multitask), solubility (single task), and toxicity prediction (multitask). Statistically significant performance increases were observed for all three prediction tasks. A potentially step change improvement was observed for the QM8 challenge which also reinforced the value of the concept of stages that privilege bonded from non-bonded interactions. 

Despite the simpler input featurization, Spatial Graph Convolutions can learn an accurate mapping of protein-ligand structures to binding free energies using the same relatively low amount of data as previous expertise-driven approaches. We thus expect that as larger sets of training data become available, Spatial Graph Convolutions can become the gold standard in affinity prediction. Unfortunately, such larger, publicly available datasets are currently nonexistent. We thus call upon academic experimental scientists and/or their pharmaceutical industry counterparts to release as much high-quality protein-ligand binding affinity data as possible so the community can develop and benefit from better affinity prediction models.

Due to the field's immense practical applications, our algorithms must prioritize realizable results over incremental improvements on somewhat arbitrary benchmarks. We thus also present a new benchmark score and accompanying cross-validation procedure.
The latter draws on agglomerative clustering of sequence and structural similarity to construct challenging train-test splits.
Using this proposed cross-validation schema, on sequence-based splitting (Table~\ref{sequence_split}) we observe in the Pearson correlation column that RF-score exceeds X-Score, and X-Score statistically ties Spatial Graph Convolutions.
For structure-based splitting (Table~\ref{structure_split}) we observe that Spatial Graph Convolution both RF-Score and X-Score in the Pearson correlation column.
We highlight the Pearson correlation for consistency with the literature, but present other metrics in the Tables~\ref{sequence_split} and~\ref{structure_split} from which similar conclusions could be drawn.

This construction (i.e., choice of cross-validation schema) helps assess models with a practical test set, such as one containing newly designed compounds on previously unseen protein targets. Although standard machine learning practice draws train and test sets from the same distribution, if machine leaning is to be applied to real-world drug discovery settings it is imperative that we accurately measure a given model's capacity both to interpolate within familiar regions of chemical space as well as to generalize to its less charted territories.

\section*{Methods}

\textit{Models.}
DNNs were constructed and trained with PyTorch~\cite{paszke2017automatic}. Custom Python code was used based on RDKit~\cite{rdkit} and OEChem~\cite{oechem} with frequent use of NumPy~\cite{numpy} and SciPy~\cite{scipy}.
Networks were trained on chemical element, formal charge, hybridization, aromaticity, and the total numbers of bonds, hydrogens (total and implicit), and radical electrons.
Random forest and linear regression models (i.e.,~X-Score) were implemented using scikit-learn~\cite{sklearn}; 
all random forests models were trained with $500$ trees and $6$ features per tree, and the implementation of X-Score is described in Ref.~\onlinecite{li2017structural}.
Hyperparameters for PotentialNet models trained for binding affinity, electronic properties, toxicity, and solubility studies are given in Table~\ref{table:hyperparameters1}; for toxicity and solubility models, only bond graph convolution layers are employed since there are no 3D coordinates provided for the associated datasets.
For these three tasks, random splitting was used for cross validation.
For the RNN sanity check of the ligand binding task, the best-performing LSTM sanity check was constructed with 5 layers, a hidden size of 32, 10 classes, and a learning rate of 3.45e-4.

\textit{Cross-validation on PDBBind 2007 core test set benchmark.} The core set was removed from the refined set sorted temporally to create the test set. Up to 8 hyperparameters were tuned through random search.
$K$-fold temporal cross validation was conducted within the train set for each hyperparameter set. For each held-out fold, validation set performance was recorded at the epoch with maximal Pearson correlation coefficient between the labeled and predicted values in the validation set. For each hyperparameter set, the validation score was the average Pearson score over the $K$ folds using the best epoch for each fold. The set with the best validation score was then used to evaluate test performance. The training set was split into $K$ temporal folds; for each fold, test set performance was recorded at the epoch with highest validation score.
All reported metrics are given
as the median with the standard deviation over K folds in parentheses. 

\textit{Cross-validation on PDBBind 2007 structure and sequence agglomerative clustering split benchmarks.}
Agglomerative clustering was performed with Ward's method~\cite{ward1963hierarchical}. Pairwise distance between PDB proteins was measured as either $1.0$ minus the sequence homology or the TMScore~\cite{zhang2004scoring}.
Within the train set, for each hyperparameter set, $K$ random splits within the train set were performed. For each held-out fold, validation set performance was recorded at the epoch with maximal Pearson correlation coefficient. The set with the best average Pearson score on the validation set was used to evaluate test set performance. The training set was again randomly split into $K$ folds; for each fold, test set performance was recorded at the epoch at which the held out performance was highest according to Pearson score. Metrics are reported as detailed above.

\section*{Safety statement}

No unexpected or unusually high safety hazards were encountered in this study.

\section*{Supporting Information}

Sequence- and structure-based agglomerative clustering cross-validation splittings for the PDBBind 2007 refined set

\section*{Acknowledgments}
We are grateful to the anonymous reviewers for their suggestions.
E.N.F.~is supported by the Blue Waters Graduate Fellowship.
Y.L., S.S., and J.Y.~acknowledge the support of the National Natural Science Foundation of China (11501306) and the Fok Ying-Tong Education Foundation (161003).
B.R.~was supported by the Fannie and John Hertz Foundation.
V.S.P.~is a consultant \& SAB member of Schrodinger, LLC and Globavir, sits on the Board of Directors of Apeel Inc, Asimov Inc, BioAge Labs, Freenome Inc, Omada Health, Patient Ping, Rigetti Computing,  and  is  a  General  Partner  at  Andreessen Horowitz. 
The Pande Group acknowledges the generous support of Dr.~Anders G.~Frøseth and Mr.~Christian Sundt for our work on machine learning.  The Pande Group is broadly supported by grants from the NIH (R01 GM062868 and U19 AI109662) as well as gift funds and contributions from Folding@home donors.

\appendix

\section{Computational complexity of network architectures} \label{app:complexity}

Here, we consider the complexity of the various neural architectures discussed in Sec.~\ref{sec:nns}, starting with the simple fully-connected setting.
Forward propagation can be understood as passing an input vector $x$ of length $n$ through $h$ matrix multiplications, each $\mathcal{O}(n^3)$, and $h$ elementwise nonlinear activation layers, each $\mathcal{O}(n)$.
Assuming $h \ll n$, this yields a total complexity of $\mathcal{O}(n^3)$.
Since back-propagation involves the same dimensions and number of layers, just in reverse order, it has the same complexity as the forward operation.

Complexity analysis for a 2D convolutional neural network, typical in computer vision tasks, is a bit more involved. An $m \times n$-dimensional filter on an $M \times N$ image yields $m \times n$ computations on $M \times N$ pixels, in total $\mathcal{O}(MNmn)$.
Using the Fast Fourier Transform for optimized processing, a single convolutional layer will be $\mathcal{O}(MN \log MN)$, the same complexity as the entire network.
Notice that these operations' costs grow exponentially with the dimension of Euclidean data---making the exploitation of symmetry far more important for 3D graph data.

The GGNN family of graph convolutional architectures includes effective optimizations to reduce complexity on graphs.
Let $d$ be the dimension of each node's internal hidden representation and $n$ be the number of nodes in the graph.
A single step of message passing for a dense graph requires $\mathcal{O}(n^2d^2)$ multiplications.
Breaking the $d$ dimensional node embeddings into $k$ different $\frac{d}{k}$ dimensional embeddings reduces this runtime to $\mathcal{O}(\frac{n^2d^2}{k})$.
As most molecules are sparse or relatively small graphs, these layers are typically $\mathcal{O}(\frac{nd^2}{k})$.
Although other optimizations exist, such as utilizing spectral representations of graphs, the models presented in this work build around this general GGNN framework with different nonlinearities and update rules.
None of these are sufficiently computationally expensive enough to alter the total runtime.

\section{Taskwise results for quantum property prediction} \label{app:qm8}

\begin{table}[h!]
    \small
    \centering
    \caption{QM8 Test Set Performances of All Tasks (Mean Absolute Error)}
    \begin{tabular}{c c c c c } 
    \hline
    \textbf{Task} & DTNN & MPNN & PotentialNet,& PotentialNet,\\
     &  &  & Single Update & Staged\\
    \hline
    \hline    
    E1 - CC2 & 0.0092 & 0.0084 & 0.0070 & \textbf{0.0068}\\
    \hline    
    E2 - CC2 & 0.0092 & 0.0091 & 0.0079 & \textbf{0.0074}\\
    \hline    
    f1 - CC2 & 0.0182 & 0.0151 & 0.0137 & \textbf{0.0134}\\
    \hline
    f2 - CC2 & 0.0377 & 0.0314 & 0.0296 & \textbf{0.0285}\\
    \hline    
    E1 - PBE0 & 0.0090 & 0.0083 & 0.0070 & \textbf{0.0067}\\
    \hline    
    E2 - PBE0 & 0.0086 & 0.0086 & 0.0074 & \textbf{0.0070}\\
    \hline    
    f1 - PBE0 & 0.0155 & 0.0123 & 0.0112 & \textbf{0.0108}\\
    \hline
    f2 - PBE0 & 0.0281 & 0.0236 & 0.0228 & \textbf{0.0215}\\
    \hline    
    E1 - CAM & 0.0086 & 0.0079 & 0.0066 & \textbf{0.0063}\\
    \hline    
    E2 - CAM & 0.0082 & 0.0082 & 0.0069 & \textbf{0.0064}\\
    \hline    
    f1 - CAM & 0.0180 & 0.0134 & 0.0123 & \textbf{0.0117}\\
    \hline
    f2 - CAM & 0.0322 & 0.0258 & 0.0245 & \textbf{0.0233}\\
    \hline
    \end{tabular}
    \label{tab:QM8_tasks}
\end{table}

In Table~\ref{tab:QM8_tasks} we have recorded the test set performances for all twelve tasks in the QM8 dataset using the MAE for a deep tensor neural network~\cite{schutt2016quantum} (DTNN), a message passing neural network~\cite{gilmer2017neural} (MPNN), and the staged and single update Spatial PotentialNet networks as in Sec.~\ref{subsubsec:potentialnet}.

\section*{References}

% \bibliography{library}


\end{document}